# MIDAS: Multi-sensorial Immersive Dynamic Autonomous System Improves Motivation of Stroke Affected Patients for Hand Rehabilitation

Fok-Chi-Seng Fok Kow, Anoop Kumar Sinha, Zhang Jin Ming, Bao Songyu, Jake Tan Jun Kang, Hong Yan Jack Jeffrey, Galina Mihaleva, Nadia Magnenat Thalmann and Yiyu Cai

*Abstract*—Majority of stroke survivors are left with poorly functioning paretic hands. Current rehabilitation devices have failed to motivate the patients enough to continue rehabilitation exercises. The objective of this project, MIDAS (Multi-sensorial Immersive Dynamic Autonomous System) is a proof of concept by using an immersive system to improve motivation of stroke patients for hand rehabilitation. MIDAS is intended for stroke patients who suffer from light to mild stroke. MIDAS is lightweight and portable. It consists of a hand exoskeleton subsystem, a Virtual Reality (VR) subsystem, and an olfactory subsystem. Altogether, MIDAS engages four out of five senses during rehabilitation. To evaluate the efficacy of MIDAS a pilot study consisting of three sessions is carried out on five stroke affected patients. Subsystems of MIDAS are added progressively in each session. The game environment, sonic effects, and scent released is carefully chosen to enhance the immersive experience. 60% of the scores of user experience are above 40 (out of 56). 96% Self Rehabilitation Motivation Scale (SRMS) rating shows that the participants are motivated to use MIDAS and 87% rating shows that MIDAS is exciting for rehabilitation. Participants experienced elevated motivation to continue stroke rehabilitation using MIDAS and no undesired side effects were reported.

## I. INTRODUCTION

Stroke rehabilitation is a long and taxing process that requires commitment and de- termination from patients. Therefore, it is important to ensure that the rehabilitation process is stimulating. One key factor is motivation, which is an essential factor that contributes to the exercise adherence and outcome of stroke rehabilitation. Researchers distinguish between intrinsic and extrinsic motivation [1]. Intrinsic motivation refers to behavior that is driven by internal rewards [2]. On the other hand, extrinsic motivation refers to behavior that is driven by external rewards, such as money, fame, grades, and praise [3,4].

Most patients have extrinsic motivation during stroke rehabilitation. It is essential to increase their intrinsic motivation, which can be enhanced by design features of the rehabilitation task that the patient is performing [5].

Many laboratories have created exoskeletons and serious games that utilizes bioengineering and robotics in autonomous or semi-autonomous systems for stroke rehabilitation. In [6], authors reported development of an electromyography (EMG) controlled serious game for rehabilitation. In this study subjects were instructed to control the cursor on computer screen with their non-dominant hand. The BRAVO hand exoskeleton reported in [7], is an EMG-driven hand exoskeleton for grasping cylindrical objects. A distinctive feature of BRAVO hand exoskeleton design is its intrinsic adaptability to patient's hand size. The BRAVO hand exoskeleton uses 2 electric actuators, 1 to cause flexion and extension of the thumb and the other for remaining fingers of the subject's hand. NREX [8], is a low-cost exoskeleton robot developed for upper limb rehabilitation from elbow to wrist. It uses two motorized axes for elbow and wrist joints. Authors claim that it can assist in reaching and drinking tasks. Another system by Hong Kong Polytechnic University [9], enables stroke affected subjects to perform hand training at different intensities. Subject's intention from the hemiplegic side is captured using surface electromyography (sEMG) signals. The exoskeleton hand robotic training then assists in hand opening or hand closing functional tasks. In [10], authors presented a wearable exoskeleton. This exoskeleton is designed as an anterior-support type device to achieve palmar extension. The bend angle data extracted using a glove with strain gauges from the non-paretic hand is transferred to the exoskeleton on the paretic hand to mimic the movements of the non-paretic hand. HERCULES, a 3 degrees of freedom (DoF) pneumatically operated exoskeleton arm upper limb rehabilitation is reported in [11]. The exoskeleton arm provides elbow flexion and extension, shoulder flexion and extension, and shoulder abduction and adduction. A semi-autonomous hybrid electroencephalography/ electrooculography (EEG/EOG) controlled whole arm exoskeleton is presented in [12]. The whole arm exoskeleton consisted of a shoulder-elbow module and wrist- hand module. The authors report that the new hybrid paradigm combining EEG and EOG is user-friendly. A limitation of most rehabilitation devices is non-portability. Authors in [13], suggest avoiding conventional materials for manufacturing that could make the robot bulky, expensive and difficult to transport.

Current efforts to increase motivation include creating video games [14,15], inventing interactive game-based rehabilitation tools [16–19], and inventing immersive escape room games [20,21]. Move-IT [22] is a game designed for upper limb stroke patients. It uses Oculus Rift Head Mounted Display (HMD) and a Leap Motion hand tracker. Authors in this study showed that subjects were pleased with the system after the session. It also motivated the subjects to perform rehabilitation exercises. A low-cost wearable device, integrated to VR environments for better rehabilitation process in patients with motor disabilities is presented in [23]. The device translates the movement of the subject's arm into the virtual jigsaw puzzle environment. A comparison between VR technology and conventional therapy for physical rehabilitation among patients with neurological deficits is presented in [24]. In this study, the patients were divided into two groups. One group used a VR instrument called Medical Interactive Rehabilitation Assistant (MIRA) for rehabilitation, and the other group used traditional therapy. The intrinsic motivation of the patients was assessed by a 22-item task evaluation questionnaire. It was reported that the VR session had a constructive influence on patients and empowered the patient to perform tasks autonomously. On the other hand, conventional exercises were boring, which decreased the motivation of patients for continuing treatment. In [25], authors established that integrating research in game design, motor learning, and neurophysiology changes with rehabilitation science can greatly improve the quality of rehabilitation.

Past research approaches include engaging multiple senses with tactile, visual, and sometimes audio feedback, but none offers an immersive experience. Tapping on the physiological phenomenon of neural plasticity and psychological reinforcement, MIDAS engages 4 of our 5 primary senses: tactility, visual, auditory, and olfactory. In this paper, we present the design and development of MIDAS. MIDAS is intended to be a portable immersive system inclusive of an exoskeleton, a smell releasing device, and a VR game that will help stroke patients feel more motivated during a hand rehabilitation session. MIDAS uses EMG signals taken from the subject's forearm to predict their intention before activating physical assistance (tactile) in the opening or closing of the fingers of their hand. To evaluate the performance of MIDAS, a pilot study is done using established tools to quantify the user experience, comfort, pleasure, and motivation of the subjects in using MIDAS for rehabilitation exercises.

The rest of the paper is organized as follows: Section II focuses on system description of MIDAS. Section III presents the protocols and tools for pilot study of MIDAS. Section IV presents our method of data collection for pilot study of MIDAS. In section V, we report the results of the pilot study. Finally, section VI concludes the paper.

## II. OVERVIEW OF MECHATRONIC DESIGN OF MIDAS

MIDAS consists of 3 subsystems, each of which engages one or more senses. The first subsystem is the hand exoskeleton that engages the patient's sense of touch through the opening of the hand. This function is achieved by capturing the user's intention to open their hand using an EMG signal sensor module placed on their forearm muscle belly. The second subsystem is the VR subsystem consisting of an VR headset (Oculus Quest 2), a VR game, and a VR-controller device. This subsystem engages the user's sense of sight and hearing. The last subsystem is the olfactory subsystem consisting of a face shield-cum-olfactory device and a circuit board. This subsystem engages the sense of smell. An overview of the design specifications of MIDAS is presented in Table I.

TABLE I
DESIGN SPECIFICATIONS OF MIDAS

| | |
|---|---|
| **Hand exoskeleton subsystem** | |
| **Weight** | 748 g |
| **Force and motion transmission system** | Linkage mechanism |
| **Material** | PLA 3D printed rigid parts and Smooth-on Dragon Skin 20 silicon soft pads |
| **Actuator** | One MG-996R servomotor (Operating Voltage = +5V, Stall Torque = 9.4 kg/cm, Operating speed = 0.17 s/60°, Rotation = 0°-180°, Weight = 55gm) to cause flexion and extension of subject's fingers. |
| **Securing method** | Cord-lock design for securing on the fingers and double-sided velcro straps for securing on the forearm and wrist |
| **EMG sensors** | Surface EMG sensors |
| **VR subsystem** | |
| **VR headset** | Oculus Quest 2 |
| **VR handheld control device** | Provided with Oculus Quest 2 |
| **VR Game** | Number of stages = 5, duration of each stage = 3 min, game environment scenes = natural location views for outdoor environment and earthly colour palette for indoor environment, sonic effects = Cheerful music, bird chirps, sound of mouse character, and pentatonic scales-high-pitched chimes |
| **Olfactory subsystem** | |
| **Covid-19 preventive face shield** | Curved PET frame |
| **Scent** | Mixture of lemon and bergamot scent |
| **Mechanism of scent release** | Piezoelectric transducer vibrating at ultrasonic frequency breaks the liquid scent into fine mist |
| **Control System** | |
| **Microcontroller** | Adafruit Feather nRF52832 |
| **Graphic display** | SSD1306 128x32OLED |
| **Voltage step up module** | MT3608 DC-DC step up module |
| **Power supply unit** | 2 X 3.7 V rechargeable Li-ion Panasonic NCR18650B batteries |

*A. The MIDAS Hand Exoskeleton Subsystem*

The exoskeleton is the primary subsystem of the MIDAS as it not only performs the rehabilitation task but also acts as the central network that controls the other subsystems. It compromises of two modules: the exoskeleton itself and sEMG sensors.

1) **The MIDAS Hand Exoskeleton Subsystem**

   While the essential requirement of an exoskeleton is to be able to actuate the stroke patient's hand with a significant range of motion, works of literature and user feedback research further suggest that the device must be safe, comfortable, lightweight, easy to wear, and small [26]. These are necessary to deliver an unobtrusive user experience and to improve the patient's motivation to use the device. These design objectives were kept in mind while proposing the design of MIDAS exoskeleton. They were achieved by complying with anthropomorphic design, appropriate sizing of the components, and proper material selection for each component of MIDAS exoskeleton.

   The MIDAS exoskeleton is shown in Fig. 1. The exoskeleton uses a linkage mechanism to provide flexion and extension of the patient's fingers. The rotary motion of a servo motor (MG-996R) is translated into a shaft's linear motion, which is then split four ways to actuate the mechanism on each finger. The rotary motion introduces interpolation into the linear movement, reducing sudden start and stop, allowing MIDAS to be less obtrusive while aiding the patient. To account for the difference in finger phalanx lengths and bone structures of each patient, the links connecting the shaft to the finger mechanism have been designed in such a way to allow two degrees of freedom (DoF). This prevents the device from moving the patient's fingers in an unnatural way and reduces the mechanical parts' stress, resulting in safe usage. In order to make MIDAS hand exoskeleton compact and lightweight, we choose to actuate the fore, middle, ring, and little finger by using one servomotor. This makes the MIDAS hand exoskeleton compact and lightweight. In order to facilitate quick prototyping of the exoskeleton hand, which is essential to the development of the MIDAS, 3D printing with Polylactic acid (PLA) is preferred. PLA has a good strength-to-weight ratio, hence allowing for a design that is lightweight and compact. The total weight for the exoskeleton is around 748 grams. Casted silicone pads are provided onto the parts of the device in contact with the patient's skin. These parts are namely the finger nodes, hand nodes, and wrist nodes. The soft material protects the patient's skin from the rough surface of the 3D printed parts and prolongs the duration the device can be worn. Silicone pads fabricated from Smooth-On Dragon-Skin 20 silicone is used as it is skin-safe and easily produced. Finally, attachments are provided to secure the exoskeleton on the patient's hand. For the wrist and hand nodes, a double-sided Velcro strap design that wraps around the forearm and hand is used. A tab has been stitched at the end of each Velcro strap to prevent it from coming off the lug. Providing a tab also eliminates the need to slide the straps through the lug and allows a faster and easier securing process. For the finger nodes, a cord-lock design, explicitly developed for MIDAS, is used. The cord is tightened around the finger by using the cord-locks.

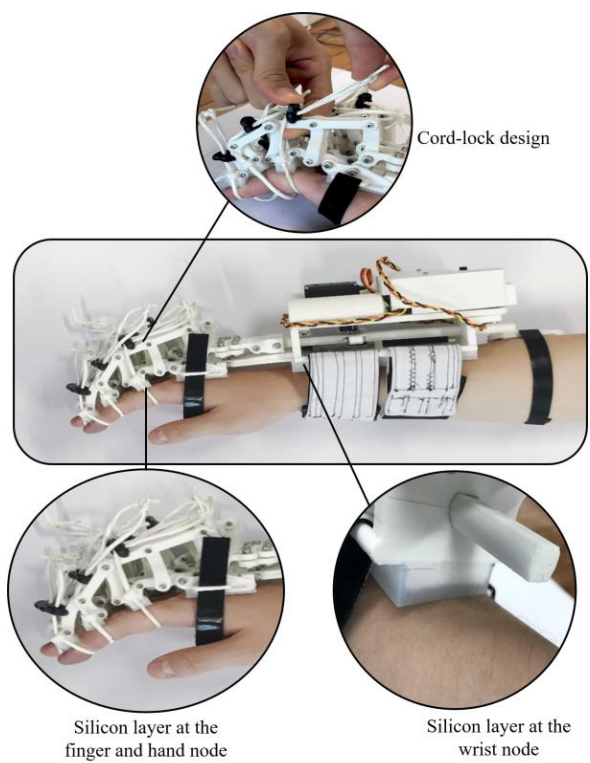

**Fig. 1.** MIDAS Exoskeleton.

2) **The EMG Sensor Module**

   In order to capture the patient's intention of hand movement, non-invasive Myoware sEMG sensor electrodes are used. The three electrodes of the sEMG sensors are placed on the forearm of the patient. The two electrodes of the sMG sensor are placed at the middle and end of a muscle belly group. The third electrode, which is the reference electrode, is placed on an inactive muscle group. The module has a duo mode, allowing the user to switch back and forth between a flexion mode and an extension mode, targeting two different muscle groups depending on the user's stroke condition. In extension mode for users with clenched hands, the sensor electrodes are placed on the lateral flexor digitorum superficialis, while in flexion mode for motor-impaired patients, the sensor electrodes are placed on the medial flexor digitorum superficialis. These muscle groups were chosen due to their functionality of gripping and releasing.

   The sEMG sensor signals are sampled at a baud rate of 115200. In order to actively filter the noise from signals, a standard moving average algorithm is then applied, in which it goes through an array of specified 50 units to derive the real-time signals.

*B. The MIDAS VR Subsystem*

The VR subsystem of MIDAS includes an Oculus Quest 2 headset, a handheld VR controller device, and a VR game. The VR subsystem is intended to engage user's sense of sight and hearing. The environments in the game, indoor and outdoor, have been designed to allow the patient to feel relax and calm while playing the game. To accomplish this, natural locations such as the forest and the beach were chosen for the outdoor environments, while an earthy color palette was adopted for indoor environments to emulate nature. In line with the environment design direction, a toon shader was applied to all models to keep the game fun, simple, and casual.

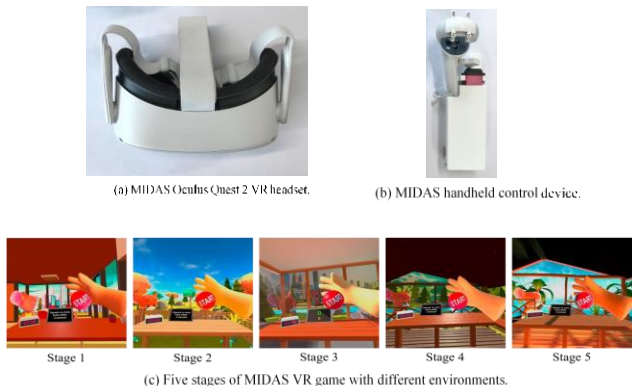

**Fig. 2.** VR subsystem of MIDAS.

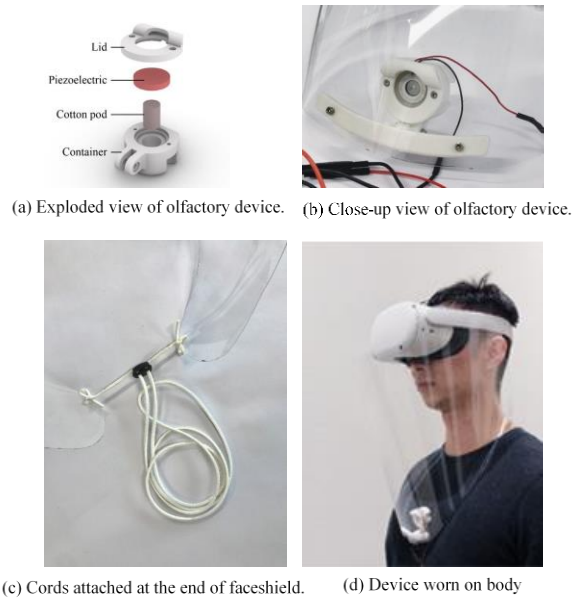

**Fig. 3.** Olfactory subsystem of MIDAS.

The game consists of five stages with different environments. Each stage in the game has a duration of 3 minutes. Fig. 2 shows the MIDAS VR headset, handheld control device, and the five stages of the game. Patients will attempt to squeeze their physical hand and through the other networking systems, allows for the digital VR hand to also squeeze. This parallax motion happens in real time and is sampled by each device at 115200 baud rate and communicated via Bluetooth. This allows for patients to associate physical motor nerve movements of their parable hand to not just haptic response with the exoskeleton, but also create motor imagery associations [27].

The mechanisms of the game-play designed to work for both types of stroke patients, i.e., patients with clenched hands (patients show flexion finger contractures) and patients who are stroke affected, but fingers of the hand do not curl into themselves. The mechanism of the game-play to score is by squeezing the lemons. It is based on the duration of both the contraction and extension of the patient's fingers. The number of squeezes increase as the game progresses. In each stage as the cups are filled with lemonade, they change to become bigger variations of cups. Patients can visually associate their progress to familiar household objects and their real-world properties. This association allows for patients, who would typically not be exposed to VR technologies, to easily integrate into this reality and accept the environment and situation.

Game sound is also an essential feature for immersive experience. Game sound, referring to all the sonic aspects of a game-ambient sounds, dialog, discrete sound effects, background music and interface sounds, has long been established as an indispensable part of modern video game [28,29]. It moves the narrative forward, convey emotions and enrich the experience of the player. Game music has the potential to be much more than a passive element of the background. By tightly linking the content of game play and background music, the player becomes much more immersed in the gaming experience. To enhance game immersion and elevate motivation of participants to keep using our device, different layers of game sounds were incorporated into the game, including cheerful background music, bird chirping, sound of a mouse character, the sound of squeezing lemon and pentatonic scales-high-pitched chimes after completing a stage.

*C. The MIDAS Olfactory Subsystem*

Smell plays a vital role in the psychophysiological effects of stress, mood and working capacity. Studies show that the inhalation of smells affects brain function via receptors in the central nervous system [30,31]. Moreover, many studies have suggested that olfactory stimulation have immediate effects on physiological aspects of humans, such as muscle tension, pupil dilation, blood pressure, pulse rate, skin temperature, and brain activity [32].

Due to the positive psychophysiological effects of smell on humans, a smell component is integrated into MIDAS for mood alteration, stress reduction, and enhancement of immersive experience of VR game. A mixture of lemon and bergamot is used to simulate the smell of lemon in the VR game. The inhalation of bergamot essential oil creates a relaxed mental state [33] and relieves work-related stress [34]. The lemon smell, on the other hand, increases theta waves in the brain [35], produces sedative effects [36], and exerts positive effects on anxiety and self-esteem [37].

The olfactory subsystem comprises two modules: a face shield-cum-olfactory device and a circuit board that connects the latter to the processing unit (see Fig. 3). This subsystem works in conjunction with the exoskeleton and the VR subsystem by diffusing a lemon and bergamot scent when activated. In line with keeping a safe operation during Covid-19, the face shield is included, as patients will have their face masks removed to smell the scent. The Fig. 3 (a) shows the exploded view of the main olfactory device. It consists of a 3D printed lid and container holding in place a piezoelectric transducer and a cotton pod soaked in lemon and bergamot scent liquid. Alternating current (AC) voltage makes the piezoelectric transducer vibrate at a very high frequency and generates ultrasonic sound. The ultrasonic sound breaks the scented liquid into a fine mist, which is instantaneously distributed in the face shield. As for the face shield, a conventional Polyethylene Terephthalate (PET) frame is used. A cord on each end of the sheet is provided. When it is pulled, the PET frame forms curve. To wear the device, the cords are passed through a cord lock to create a loop, which can then be put around the neck.

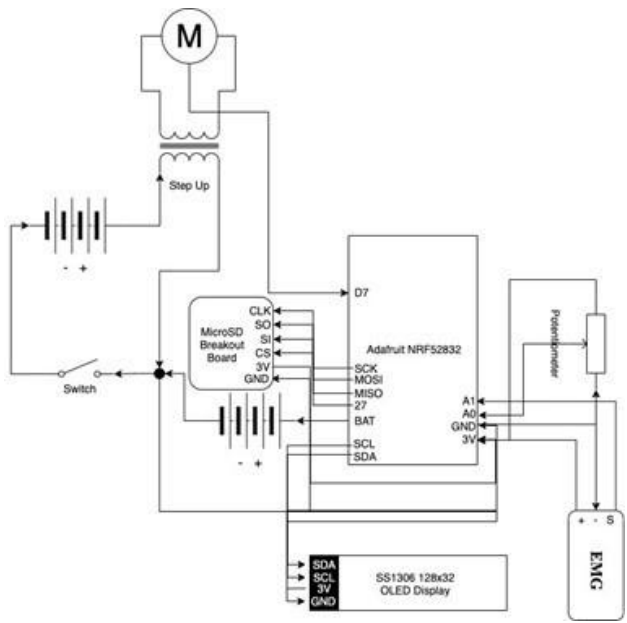

**Fig. 4.** Circuitry behind MIDAS for communication between the EMG sensors and exoskeleton actuator.

### D. Control Scheme

The MIDAS circuit for communication between hand exoskeleton and EMG sensors is shown in Fig. 4. The chosen interface between the 3D printed exoskeleton and the user was commercially available microcontroller circuit boards. In our case, the Adafruit Feather system was chosen due to its compact size and built-in functionalities. The circuitry runs on the Adafruit Feather nRF52832 microcontroller, SSD1306 128x32OLED graphic display, MG996 servo motor, MT3608 DC-DC voltage step up module, Adafruit MicroSD SPI card breakout board, rotary potentiometer, an electrical switch and 2 sets of 2 Panasonic NCR18650B batteries. The overall network diagram showing channel for communication between different subsystems of MIDAS is shown in Fig. 5. The MIDAS exoskeleton circuitry receives its commands from the EMG sensor module and relays this information to the microprocessor where it will process the data through cleaning up the noise and applying an averaging algorithm, saving the data on the MicroSD card, sending the PWM data to MG996 servo to actuate the exoskeleton and, sending the real time data acquired to the VR subsystem and Olfactory subsystem. The microchip-based microcontroller (model Adafruit Feather NRF52832) is ARM based, contains the latest Bluetooth technologies and extremely energy efficient, all thanks to its single chip architecture. By utilizing the latest nRF52832 MCU, it not only allows for faster and more efficient data transfer between Bluetooth peripherals, but also allows for the microcontroller to behave as a central connection between other multiple Bluetooth peripherals. This central hub mode is still in its early stages of development but has since made this project possible with almost no more external networking peripherals.

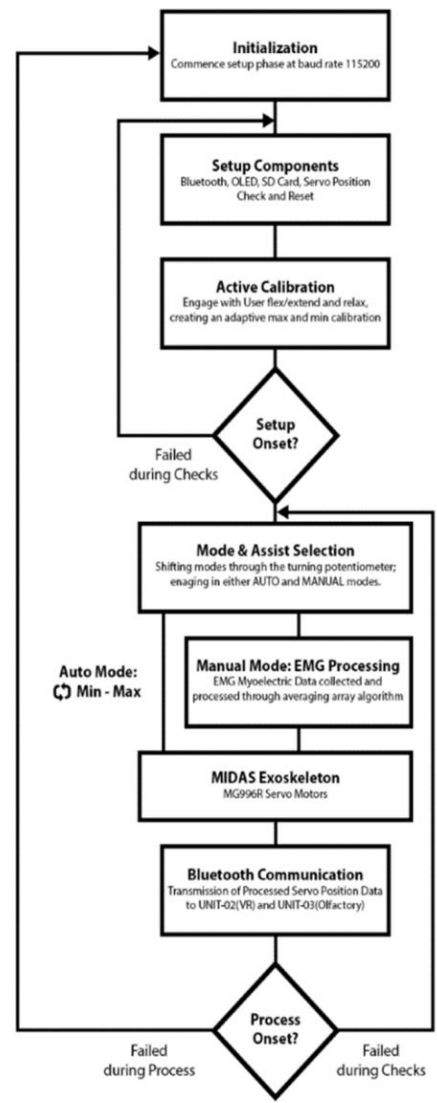

**Fig. 5.** Flow chart of information processing in MIDAS system.

### III. PROTOCOLS AND TOOLS FOR PILOT STUDY OF MIDAS

Since this is a preliminary test, only 5 post-stroke patients were recruited to test the functionality and applicability of the device. The post-stroke patients were chosen according to the following criteria:

(1) age from 21 to 65
(2) diagnosed with light to mild stroke and still recovering
(3) passed the rapid cognitive screening (rapid cognitive screen score > 7).

### A. Recruitment

The research team put up posters at local stroke rehabilitation organizations. Interested participants contacted the research team for participation. Consent was taken after the researchers conducted screening tests and briefly explained the study objectives to patients. The process of data collection was carried out from the end of February 2021 to the end of March 2021 at the participant's house or a place agreed upon by both the researchers and the participant.

TABLE II
DEMOGRAPHIC DETAILS OF STROKE PATIENTS

| Subject number | Gender | Age (years) | Level of education | Marital status | Severity of stroke | Impaired side | Time since stroke onset (months) |
|---|---|---|---|---|---|---|---|
| 1 | Male | 56 | Between secondary and tertiary | Married | Mild | Left | 38 |
| 2 | Female | 52 | Tertiary | Single | Mild | Left | 36 |
| 3 | Female | 27 | Tertiary | Married | Light | Right | 17 |
| 4 | Male | 61 | Tertiary | Married | Mild | Left | 162 |
| 5 | Male | 46 | Tertiary | Single | Light/mild | Right | 114 |

Out of the 5 participants recruited, 3 were male. 3 were married and living with their partner. 4 participants received tertiary education. 3 of them were impaired on the left side and suffered from mild stroke. The youngest participant was 27 years old; the oldest was 61. The time since stroke onset ranged from 17 months to 162 months. The demographic data relating to the stroke patients are mentioned in Table II.

*B. Tools of Data Collection*

1) **Structured Interviewing Questionnaire Form**
   The questionnaire includes socio-demographic characteristics (i.e., age, sex, marital status, educational level and occupation) and medical diagnosis.

2) **Motor Assessment Scale (MAS)**
   To evaluate motor function in recovering stroke patients, the MAS presented in [38], is used in this study. The MAS uses tasks related to activities of daily living to measure the full range of functional motor performance in stroke survivors. These consist of 9 items which are supine to side lying, supine to sitting over the side of the bed, balanced sitting, sitting to standing, walking, upper-arm function, hand movements, advanced hand activities, and general tonus. Items (with the exception of the general tonus item) are assessed using a 7-point scale (0 to 6). A score of 6 indicates optimal motor behavior. Item scores (with the exception of the general tonus item) are summed to provide an overall score (out of 48 points). For the general tonus item, the score is based on continuous observations throughout the assessment. A score of 4 on this item indicates a consistently normal response, a score less than 4 indicates persistent hypertonus, and a score greater than 4 indicates various degrees of hypotonus.

3) **Montreal Cognitive Assessment (MoCA)**
   MoCA is developed as a rapid screening instrument for mild cognitive dysfunction [39]. It assesses different cognitive domains: attention and concentration, executive functions, memory, language, visuoconstructional skills, conceptual thinking, calculations, and orientation. It consists of 7 parts: visuospatial/executive, naming, memory, attention, language, abstraction, delayed recall, and orientation.
   The maximum score for this tool is 30 points. A final total score of 26 and above is considered normal. One point is added for an individual who has 12 years or fewer of formal education.

4) **WHOQOL-BREF**

TABLE III
USER EXPERIENCE SCORING SYSTEM

| | Score | | | | | | | |
|---|---|---|---|---|---|---|---|---|
| Obstructive | ○ | ○ | ○ | ○ | ○ | ○ | ○ | Supportive |
| Complicated | ○ | ○ | ○ | ○ | ○ | ○ | ○ | Easy |
| Inefficient | ○ | ○ | ○ | ○ | ○ | ○ | ○ | Efficient |
| Clear | ○ | ○ | ○ | ○ | ○ | ○ | ○ | Confusing |
| Boring | ○ | ○ | ○ | ○ | ○ | ○ | ○ | Exciting |
| Uninteresting | ○ | ○ | ○ | ○ | ○ | ○ | ○ | Interesting |
| Conventional | ○ | ○ | ○ | ○ | ○ | ○ | ○ | Inventive |
| Usual | ○ | ○ | ○ | ○ | ○ | ○ | ○ | Leading edge |

It is an abbreviated version of the WHOQOL-100, which was developed by the WHO-QOL Group in 1994 [40]. It is to cross-culturally assess quality of life. It consists of 4 domains: physical health, psychological, social relationships, and environment. It is a generic instrument to assess the positive and negative aspects of quality of life. The questions relating to the 4 facets i.e., physical pain and discomfort experienced, psychological positive or negative feelings about life, social relationships such as companionship, love, and support desired in the personal relationship, and environmental such as patient's financial resources to live a healthy and comfortable life were asked. The raw scores from the questions of these domains were then assigned.

5) **User Experience Questionnaire**
   To evaluate the user experience of MIDAS in this study, an adapted version of the original user experience questionnaire reported in [41], is used. It contains 8 items, namely obstructive or supportive, complicated or easy, inefficient or efficient, clear or confusing, boring or exciting, non-interesting or interesting, conventional or inventive, and usual or leading edge. Each item is rated on a 7-point Likert scale. Table III shows the scoring system for user experience questionnaire.

6) **Stroke Rehabilitation Motivation Scale (SRMS)**
   The original SRMS, which consisted of 28 questions, was presented in [42]. It is used to measure internal and external factors on motivation among stroke patients to re-habilitate. A shorter version was also developed by the original creators of the scale. It comprises 7 questions which are correlated with the original 28 questions from the original SRMS. For this study, the shorter version is adapted, and the 7 questions are:
   a) If you have choices, do you feel like choosing to do rehabilitation using our device?
   b) Do you feel like your motivation for participating in rehabilitation has gotten higher after trying our device?

TABLE IV
SAM SCORING SYSTEM

| Scale | Valence rating | Arousal rating | Dominance rating |
|---|---|---|---|
| 5 | Pleasant | Excited | Dependent |
| 4 | Pleased | Wide-awake | Powerlessness |
| 3 | Neutral | Neutral | Neutral |
| 2 | Unsatisfied | Dull | Powerful |
| 1 | Unpleasant | Calm | Independent |

    c) Do you feel the device could be useful in helping you to exercise your hand?
    d) Do you want to participate in rehabilitation with our device if you see other stroke patients getting better through rehabilitation?
    e) Would you perform rehabilitation with our device even if no one is enforcing it on you?
    f) Do you find participating in rehabilitation with our device exciting?
    g) Does doing rehabilitation with our device help you feel like you are achieving something?
    h) Each item is rated in a 5-point Likert scale. A score of 1 = completely disagree to a score of 5 = completely agree. The score ranges from 7 to 35. A higher score means higher motivation.

7) **Self-Assessment Manikin (SAM)**
SAM has been devised to assess the emotional response of a person to an object or event. Although there is ongoing debate regarding what defining features should be considered to capture a person's emotions, authors in [43] have presented three dimensions namely pleasure, arousal, and dominance associated with a person's effective reaction to a stimulus to measure a person's emotions. In this study, a 5-point scale is used to measure a person's emotional response to MIDAS. Table IV shows the scoring system for this tool.

## IV. METHOD OF DATA COLLECTION

The data for the pilot study of MIDAS was collected in 3 sessions. The three subsystems of MIDAS were included in the study progressively in each session, one by one. In the 1st session, the patients were asked to wear the hand exoskeleton with the Myoware sEMG Sensors. The patient's response to using the device was then recorded. The second session included recording the patient's response when the hand exoskeleton along with the VR subsystem is used. In the third session, all the subsystems of MIDAS, i.e., the hand exoskeleton, the VR, and the olfactory subsystem, were used, and the patient's response was recorded. This was done to evaluate the performance of MIDAS in engaging one or more senses of the patients in stages and its acceptability by the patients. In the 2nd and 3rd session, the patients were asked to play the VR game. Fig. 6 (a) shows the VR game and Fig. 6 (b) - (d) shows the patients using MIDAS subsystems in the 3 different sessions.

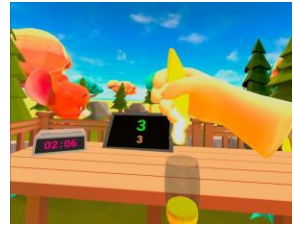
(a) The VR game showing the lemon squeezed to fill a cup.

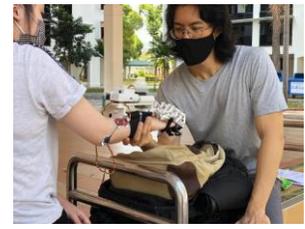
(b) Patient using hand exoskeleton during session 1.

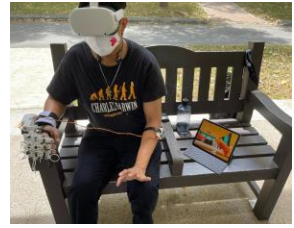
(c) Patient using hand exoskeleton and VR during session 2.

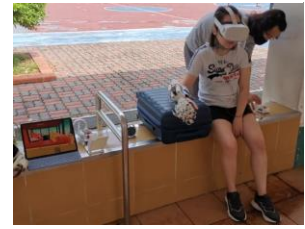
(d) Patient using hand exoskeleton, VR, and olfactory device during session 3.

**Fig. 6.** VR game environment and pilot study of MIDAS during the three sessions.

## V. RESULTS

In this section, the results of the pilot study of MIDAS are presented.

### A. MAS, MoCA, and QOL

As shown from the MAS (see Fig. 7 (a)), the motor functions of subjects vary greatly, ranging from 21 to 54. There is no significant difference of motor functions between and first and last sessions. MoCA results in Fig. 7 (b) indicate that all participants are cognitively intact (with MoCA score greater than 26). The quality of life of participants also varies greatly (see Fig. 7 (c)). The lowest and highest scores are: 38/94 (physical), 50/88 (psychological), 50/94 (social relationship), 56/94 (environmental).

### B. User Experience Feedback

Fig. 8 shows the overall user experience scores given by each subject. Table V shows the ratings provided the subjects during the 1st and 3rd session for each category of the user experience questionnaire. The user experience feedbacks of the 3rd session are higher than those of the 1st session for all subjects. 60% of the scores are above 40 (out of 56). Most subjects gave higher ratings on the 3rd session than they did on the 1st session. 80% of participants found the device complicated to use during the 1st session. However, during the 3rd session only 1 participant reported that the device is complicated. This difference may have occurred because they were more familiar with the device during the last sessions. During the 1st session only 1 participant completely agreed that the device is inventive. However, during the 3rd session 60% of the participants completely agree that device is inventive. This shift may have occurred because of subsequent addition of the VR and olfactory subsystem of MIDAS during the 2nd and 3rd session,

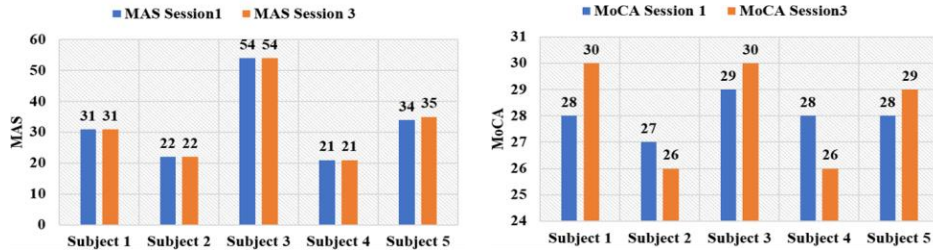

(a) MAS score of subjects for session 1 and session 3.  (b) MoCA score of subjects for session 1 and session 3.

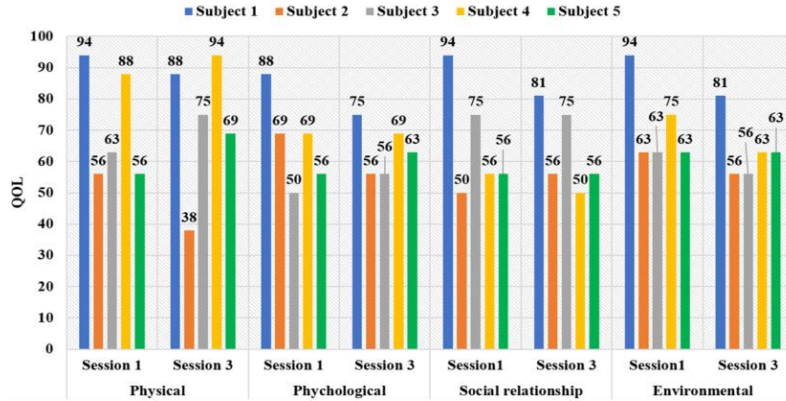

(c) QOL score of subjects for session 1 and session 3.

**Fig. 7.** MAS, MoCA, and QOL scores of all subjects considered in this pilot study for session 1 and session 3.

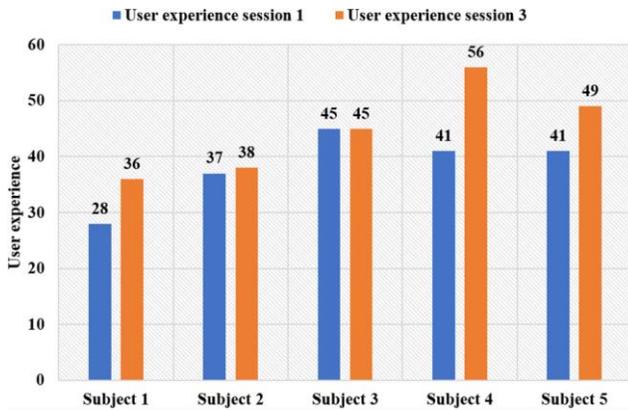

**Fig. 8.** User experience feedback of subjects for session 1 and session 3.

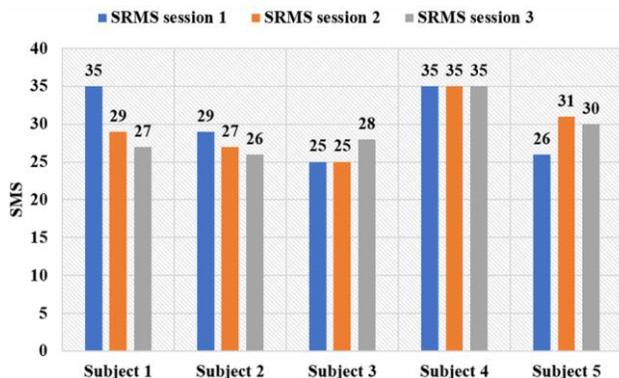

**Fig. 9.** SRMS score of subjects for all sessions.

respectively. The ratings for system being efficient also increased on the 3rd session.

### C. Self-rehabilitation Motivation of the Subjects

Fig. 9 shows the SRMS scores of the subjects for all sessions. Self-assessed motivation varied greatly, depending on each participant. Across all 3 sessions, 2 participants gave decreasing scores, 2 participants gave increasing scores, 1 participant gave full marks for all session. This indicates that 3 participants out of five were motivated to use MIDAS. Table VI shows the 7 questions adopted for this study. The data in the table indicates the SRMS Likert response/percentage of subjects in each category for each question. It can be seen from the Table VI that all participants would choose our device to do rehabilitation (rated as somewhat agree or completely agree). 73% of ratings showed that participant's motivation had gotten higher after trying our device. 80% of ratings show that the participants find MIDAS useful in performing exercises for hand rehabilitation. 94% rating show that the participants are willing to use MIDAS without anybody enforcing it on them. 87% rating shows that the MIDAS is exciting for rehabilitation and 60% of ratings showed that participants felt they achieved something with our device.

### D. SAM

Fig. 10 shows the variation in emotional response of the subjects towards using MIDAS during the three sessions. The results indicate that there was no significant change in pleasure ratings after the device trial during the 1st session. During the 2nd session, 2 subjects reported increased pleasure after the device usage. However, during the 3rd session, 1 subject reported decreased pleasure, 2 subjects reported increased pleasure, remaining 2 subjects reported no change in pleasure. Arousal ratings also varied during the three sessions. 1 subject reported increase in arousal during the 1st session while the remaining subjects showed no significant increase in arousal rating. During the 2nd session, 2 subject reported increase in arousal. While during the 3rd session, 1 subject reported increased arousal. Although dominance ratings did not change at all during the 3rd session, 2 subjects in 1st and 2nd session reported increased dominance after the device trial.

TABLE V
RESPONSE OF THE SUBJECTS TO THE USER EXPERIENCE QUESTIONNAIRE

| Item | Completely disagree 1st/3rd | Disagree 1st/3rd | Slightly disagree 1st/3rd | Neither agree nor disagree 1st/3rd | Slightly agree 1st/3rd | Agree 1st/3rd | Completely agree 1st/3rd |
|---|---|---|---|---|---|---|---|
| Supportive | 0/0 | 0/0 | 0/0 | 2/0 | 1/1 | 2/3 | 0/1 |
| Easy | 0/0 | 0/0 | 4/1 | 0/1 | 0/0 | 1/2 | 0/1 |
| Efficient | 0/0 | 0/0 | 2/1 | 1/1 | 2/2 | 0/0 | 0/1 |
| Clear | 0/0 | 0/0 | 1/0 | 1/1 | 2/1 | 1/2 | 0/1 |
| Exciting | 0/0 | 0/0 | 0/1 | 2/1 | 0/1 | 2/0 | 1/2 |
| Interesting | 0/0 | 0/0 | 0/0 | 1/1 | 0/2 | 2/0 | 2/2 |
| Inventive | 1/0 | 0/0 | 1/0 | 0/0 | 0/0 | 2/2 | 1/3 |
| Leading edge | 0/0 | 1/0 | 0/0 | 0/1 | 1/0 | 2/3 | 1/1 |

TABLE VI
RESPONSE OF THE SUBJECTS TO THE SRMS QUESTIONNAIRE

| Questions | Completely disagree | Disagree | Neither agree nor disagree | Somewhat agree | completely agree |
|---|---|---|---|---|---|
| If you have choices, do you feel like choosing to do rehabilitation using our device? | 0/0 | 0/0 | 0/0 | 9/60 | 6/40 |
| Do you feel like your motivation for participating in rehabilitation has gotten higher after trying our device? | 0/0 | 0/0 | 4/27 | 6/40 | 5/33 |
| Do you feel the device could be useful in helping you to exercise your hand? | 0/0 | 0/0 | 3/20 | 5/30 | 7/50 |
| Do you want to participate in rehabilitation with our device if you see other stroke patients getting better through rehabilitation? | 0/0 | 0/0 | 1/7 | 8/53 | 6/40 |
| Would you perform rehabilitation with our device even if no one is enforcing it on you? | 0/0 | 0/0 | 1/7 | 7/47 | 7/47 |
| Do you find participating in rehabilitation with our device exciting? | 0/0 | 0/0 | 2/13 | 6/40 | 7/47 |
| Does doing rehabilitation with our device help you feel like you are achieving something? | 1/7 | 0/0 | 5/33 | 5/33 | 4/27 |

## VI. CONCLUSION

In this paper, a Multi-sensorial Immersive Dynamic Autonomous System called MIDAS is presented. In order to influence user's experience and behavior, MIDAS increases the number of stimuli for rehabilitation by engaging four out of five senses. It consists of three subsystems, i.e., a hand exoskeleton, a VR headset with a handheld controller, and an olfactory device. Since most components of MIDAS are 3D printed, they can be easily reproduced in any research laboratory for further research. MIDAS is lightweight and portable. This allowed us to meet the subjects during each session at their house or a place agreed upon and perform the pilot study for this research.

Every component in the subsystems of MIDAS is carefully designed to provide maximum immersive experience and pleasure to the user. The MIDAS hand exoskeleton is comfortable and easy to wear. The natural location views in the VR game environment with cheerful music, bird chirps, and a mouse character keep the user engaged and motivated. The olfactory subsystem releases lemon and bergamot scent each time the subject squeezes the lemon in the game. This keeps the subject relaxed and stable.

MIDAS is a novel system that provides a more immersive experience for hand rehabilitation of stroke-affected patients. The user experience study of MIDAS indicates that it is easy for users to familiarize themselves with it. The subjects also reported that the device is inventive and effective in hand rehabilitation. Most subjects gave higher ratings at the 3rd session than they did at the 1st session. SRMS ratings of MIDAS indicate that the device increased the subject's motivation to use it for hand rehabilitation. The subjects find it exciting and are willing to use MIDAS for rehabilitation.

MIDAS sets an example for the researchers on how multiple senses of patients can be engaged during rehabilitation

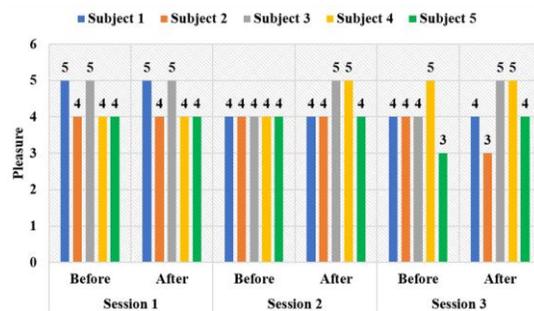
(a) Difference in pleasure response of the subjects before and after using MIDAS during the three sessions.

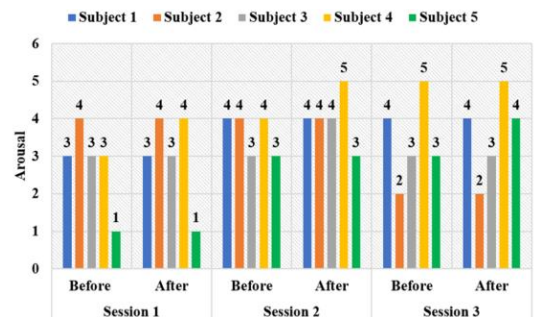
(b) Difference in arousal response of the subjects before and after using MIDAS during the three sessions.

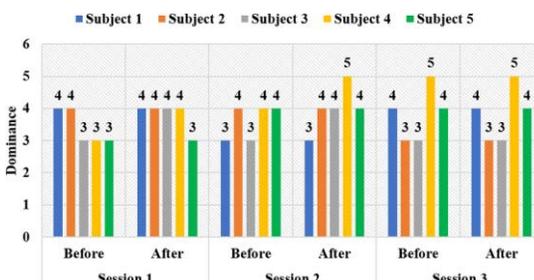
(c) Difference in dominance response of the subjects before and after using MIDAS during the three sessions.

**Fig. 10.** SAM scores of subjects for session 1, session 2 and session 3.

exercises. It provides proof of increased motivation in rehabilitation by making the rehabilitation exercise exciting and highly immersive.

Future researchers must keep in mind that needs and conditions vary for each stroke patient depending on their age, medical condition, and severity of stroke. Hardly can a device meet the needs of all stroke survivors. The first major area of improvement in MIDAS is the hand exoskeleton. High torque and power output from the motor would increase the range of movement of the patient's fingers. This would also increase the range of stroke patients who can use the system. However, for this to be possible, there are two prerequisites. The first is a higher power supply, as a stronger motor means a higher power demand. The second is a more robust material for the exoskeleton so that it can withstand higher stress. One possible material is resin. It performs well under high stress while also being cheap and fast to produce compared to other materials like Acrylonitrile Butadiene Styrene (ABS) or carbon fiber. In our current design of the olfactory subsystem, a click sound can be heard before and after the scent's release. This might be intrusive during the rehabilitation. To reduce this disturbance, silent relays could be used to trigger the scent. Different types of fruits, along with its corresponding smell, can also be used in each stage to further make the game more engaging.

Although no substantial improvements in participant's hand functions were noticed at the end of the pilot study, participants found MIDAS interesting, exciting and are supportive of it. This pilot study provides preliminary data suggesting that a device like MIDAS, which intends to engage more patients' senses and provide an immersive experience during rehabilitation exercises, is more effective than the conventional simple hand exoskeleton devices. A limitation of this study was fewer participants and a small number of sessions. Future studies need to be done with a bigger sample size, increased number of sessions, and the duration of each session

ACKNOWLEDGMENT

This research is supported by Accelerating Creativity and Excellence (ACE) grant; and Institute for Media Innovation (IMI), Nanyang Technological University, Singapore.

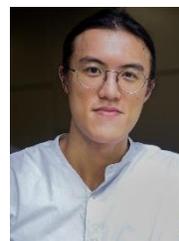

**Fok-Chi-Seng Fok Kow** received his BFA in Product Design from the Nanyang Technological University (NTU). He is currently a Research Assistant at the NTU, School of Art, Design and Media exploring Singapore as a method for inclusivity. He has previously been nominated for the ADM Sustainability Award and has done multiple exhibitions. His research interests lie in parametric and sustainable design.

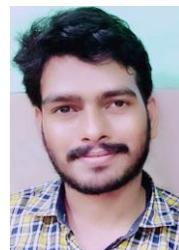

**Anoop Kumar Sinha** received B.Tech degree in mechanical engineering from Dr. A.P.J. Abdul Kalam Technical University, Lucknow, India and M.Tech degree in mechatronics engineering from Indian Institute of Technology (IIT) Patna, India. He is currently working toward the Ph.D. degree in mechanical engineering with School of Mechanical and Aerospace Engineering, Nanyang Technological University (NTU), Singapore.

He has invented 2 patents (one granted and second pending approval) in the area of biaxial stretching device and cryogenic micromachining of soft materials. His current research interest includes prostheses, sensor feedback, human robot interaction and mechatronics.

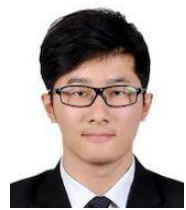

**Zhang Jinming** received his bachelor's degree in psychology from National University of Singapore in 2019. His research interests are subliminal message, hypnosis, and human perception of art.

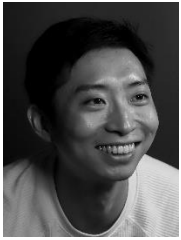

**Bao Songyu** is a kinetic media artist and prototype designer and is currently pursuing his master's in arts at Nanyang Technological University. He received his BFA Degree in Interactive Media and Minor in Entrepreneurship in 2019. Bao's work ranges from installations to kinetic sculptures with elements of robotics and artificial life. He aims to continue experimenting with different materials used in 3D printing and production to further his understanding of materials and their applications.

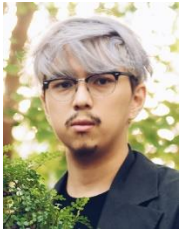

**Jake Tan Jun Kang** is a New Media Artist, Adjunct Lecturer at Temasek Polytechnic School of Design and Founder of Serial Communication, a studio that focuses on Research & Development of proprietary digital assets in the fields on Creative Technology and Extended Reality (XR). Serial Communication is 1 of the 2 selected Start-ups that was selected for an incubation program by Nanyang Technological University, School of Art, Media and Design, Singapore, where he also received his BFA degree in Interactive Media in 2020, and MiniMasters in General Management in Nanyang Business School. His R&D platforms typically lie in the physical platforms of microcontrollers, electronics, 3D printing and scanning, digitally through programming, AR/VR/MR production through game engines and intangibly through projection mapping and XR focused user interface and experience design. His artistic practice encompasses topics of intersection between nature, technology and society, where he has exhibited and was part of artistic residencies in Singapore, Austria and Germany.

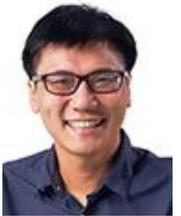

**Hong Yan Jack Jeffrey** is currently a Lecturer for the Product Design pathway at the School of Art, Design, and Media at Nanyang Technological University in Singapore. Trained in architecture, industrial design and human factors engineering, he has worked on multi-disciplinary design projects ranging from consumer product design, interaction design, exhibition design, and transportation design.

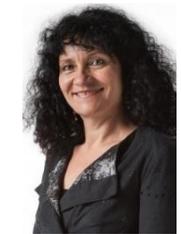

**Galina Mihaleva** is an Associate Professor at the school of Art, Design and Media at Nanyang Technological University, where she teaches Wearable Technology, Fashion and Design. Her work and research deal primarily with the dialogue between body and dress, driven by the idea of having both a physical and a psychological relationship with a garment as a responsive clothing - wearable technology.

Prior to joining NTU, Galina thought at Arizona State University for more than 15 years costume and fashion design and collaborated with world renown choreographers in USA, ASIA and Europe. She is the founder of Galina Couture in Scottsdale Arizona, where her team develops exclusive collections of one of kind designs. Her art and design work have been shown in festivals, galleries and museums across United States, Asia, Central and South America and Europe. In 2007 she was nominated for the best design award at Cooper-Hewitt Design Museum. Galina received the Rumi award in USA and the first place at the Tiffany's Paris fashion week in 2016.

Her resent interests and explorations are in the area of Fashionable Technology, which refers to the intersection of design, fashion, science, and technology.

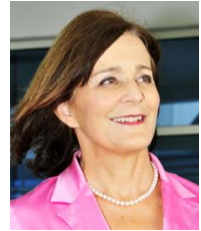

**Nadia Magnenat Thalmann** received bachelor's and master's degrees in disciplines such as psychology, biology, chemistry, and computer science, and completed her PhD in quantum physics at the University of Geneva. She has received honorary doctorates from Leibniz University of Hannover and the University of Ottawa in Canada. Since 1989, she is a Professor at the University of Geneva where she founded the interdisciplinary research group MIRALab. She is presently Director of the IMI in NTU, Singapore. Her global domain of research is Virtual Humans and Social Robots.

She revolutionized social robotics by unveiling the first social robot Nadine at NTU. She has authored dozens of books and published with her students more than 700 papers on Virtual Humans/Virtual Worlds and social Robots. She is the chair of major conferences as CGI, CASA, and has delivered more than 300 keynote addresses, some of them at global events such as the World Economic Forum in Davos.

Prof. Thalmann is Editor-in-Chief of the Visual Computer and Associate editor of many other scientific journals. She is a life member of the Swiss Academy of Engineering Sciences. She is a recipient of the prestigious Humboldt Research Award in Germany. wikipedia/Nadia_Magnenat_Thalmann.

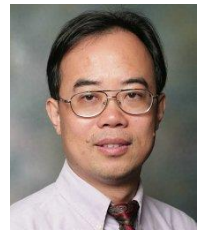

**Yiyu Cai** did his Ph.D. training in Engineering, M.Sc. training in Computer Graphics & Computer-aided Geometry Design, and B.Sc. training in Math. He is currently a tenured faculty with the School of Mechanical & Aerospace Engineering in NTU Singapore, Program Director of The Strategic Research Program of Virtual Reality and Soft Computing, Professor in Charge of The Computer-aided Engineering Labs.

He has edited 8 books and 5 journals special issues, and published over 200 papers in peer-reviewed international journals or international conferences. He has also invented or co-invented 6 patents (granted or pending approval). His research interest includes 3D based design, simulation, serious games, virtual reality, etc.